\documentclass{article}
\usepackage{spconf,amsmath,graphicx}
\usepackage[backref]{hyperref}  % Jie added
\usepackage{algorithm}
\usepackage{algorithmic}
\usepackage{booktabs}
\usepackage{amsmath} 
\usepackage[compact, center, small,up]{titlesec}

\title{HCIL: Hierarchical Class Incremental Learning for Longline Fishing Visual Monitoring}
\name{Jie Mei$^{\star}$\thanks{$^{\star}$e-mail:\{jiemei,hwang\}@uw.edu},  Suzanne Romain$^{\dagger}$, Craig Rose$^{\dagger}$, Kelsey Magrane$^{\dagger}$\thanks{$^{\dagger}$e-mail:\{suzanne.romain,craig.rose,kelsey.magrane\}
@noaa.gov}, Jenq-Neng Hwang$^{\star}$}
  
\address{$^{\star}$ Department of ECE,
University of Washington,
Seattle, WA, USA\\
$^{\dagger}$Pacific States Marine Fisheries Commission, National Oceanic and Atmospheric Administration, USA}

\begin{document}
%\ninept
%
\maketitle
\begin{abstract}
The goal of electronic monitoring of longline fishing is to visually monitor the fish catching activities on fishing vessels based on cameras, either for regulatory compliance or catch counting. The previous hierarchical classification method demonstrates efficient fish species identification of catches from longline fishing, where fishes are under severe deformation and self-occlusion during the catching process. Although the hierarchical classification mitigates the laborious efforts of human reviews by providing confidence scores in different hierarchical levels, its performance drops dramatically under the class incremental learning (CIL) scenario. A CIL system should be able to learn about more and more classes over time from a stream of data, i.e., only the training data for a small number of classes have to be present at the beginning and new classes can be added progressively. In this work, we introduce a Hierarchical Class Incremental Learning (HCIL) model, which significantly improves the state-of-the-art hierarchical classification methods under the CIL scenario.
\end{abstract}
\section{\textbf{Introduction}}
\label{sec:intro}
\textbf{Electronic Monitoring (EM) of Fisheries} Automated imagery analysis techniques have drawn increasing attention in fisheries science and industry \cite{mei2022unsupervised, guptatrends, white2006automated,mei2021absolute,huang2016chute,williams2016automated,huang2016live,wang2016closed} because they are more scalable and deployable than conventional manual survey and monitoring approach.

\begin{figure}
\centering\includegraphics[width=0.45\textwidth]{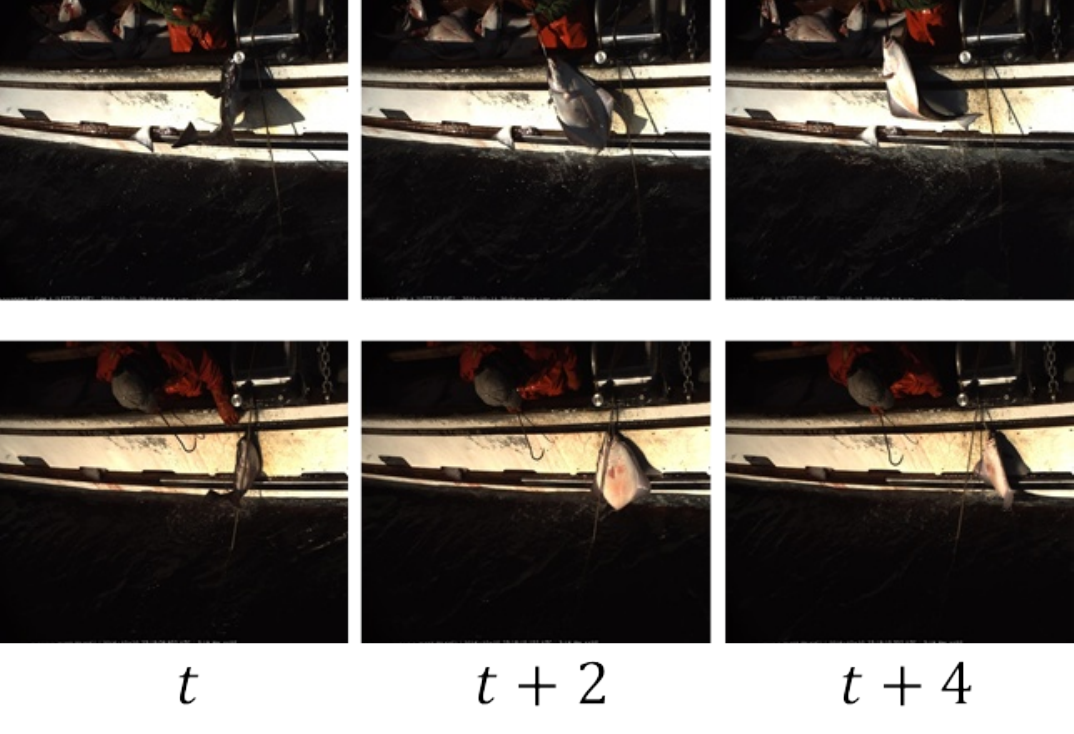}
\caption{Longline Fishing: Each row is a temporal sequence of an individual fish caught on a longline hook, as it is being pulled up from the sea and over the rail of the fishing vessel.} \label{vessel}
\end{figure}

The goal of EM is to systematically monitor fish captures using cameras on fishing vessels either for catching counting or regulatory compliance. Then fisheries managers can thus assess the count of fish caught by species and size to monitor catch quotas by vessel or fishery. Besides, managers will detect the retention of specific fish species or sizes that are illegal to be kept. Therefore, accurate detection, length measurement, and species identification are critically needed in the EM systems. In this work, our approach focuses on the species identification task for the video-based longline fishing monitoring, where fish are caught on hooks and viewed as they are pulled up from the sea and over the rail of the fishing vessel as shown in Fig.\ref{vessel}.

\textbf{Hierarchical Classification}
For the species identification task in the EM systems, a hierarchical classifier has more practical use for the fisheries managers than a flat classifier because it can predict coarse-level groups and fine-level species at the same time. Both levels' predictions are useful for fisheries managers to evaluate the status of fish stocks. 

The previous hierarchical fish classification work~\cite{Mei2020VideobasedHS} enforces the hierarchical data structure and introduces an efficient training and inference strategy for video-based fisheries data. With the hierarchical inference, if some input images are predicted with high confidence in one coarse-level group but with low confidence in the corresponding fine-level species, then the hierarchical model allows fisheries personnel to further assign appropriate experts to review those images and get the correct fine-level labels. 

\textbf{Class Incremental Learning}
Deep neural networks achieve remarkable performance in supervised classification tasks, but only when all the classes to be learned are available at the same time. However, real-world data are constantly acquired through time, leading to ever-changing distributions, i.e., new target fish species are added continuously. When a deep neural network model loses access to previous classes data (e.g., for privacy reasons, storage limitations, or data transfer difficulties) and can only be finetuned on new classes data, it could catastrophically forget the old classes, the so-called catastrophic forgetting problem~\cite{douillard2021dytox, rebuffi2017icarl, hou2019learning, castro2018end}.

Although the previous work~\cite{Mei2020VideobasedHS} achieves the state-of-the-art performance on hierarchical species classification task, its performance drops dramatically under the class incremental learning (CIL) scenario. A CIL system should be able to learn more and more classes over time from a stream of data, where only the training data for a small number of classes are present at the beginning and new classes can be added progressively. 

Our proposed Hierarchical Class Incremental Learning (HCIL) method can provide coarse-level prediction and fine-level species at the same time, while the system gradually acquires increasing number of training fish classes over the time. More importantly, it significantly improves the state-of-the-art hierarchical classification method under the class incremental learning scenario. 

The remaining sections of this paper are organized as follows. In Section~\ref{Related Work}, overviews of the related works in class incremental learning and hierarchical classification are provided. Section~\ref{Proposed Method} describes details of our proposed method, HCIL. The experimental results are demonstrated and discussed in Section~\ref{Experimental Results}. Finally, Section~\ref{Conclusions} gives conclusions and future work.

\begin{figure} 
\centering
\includegraphics[width=0.52\textwidth]{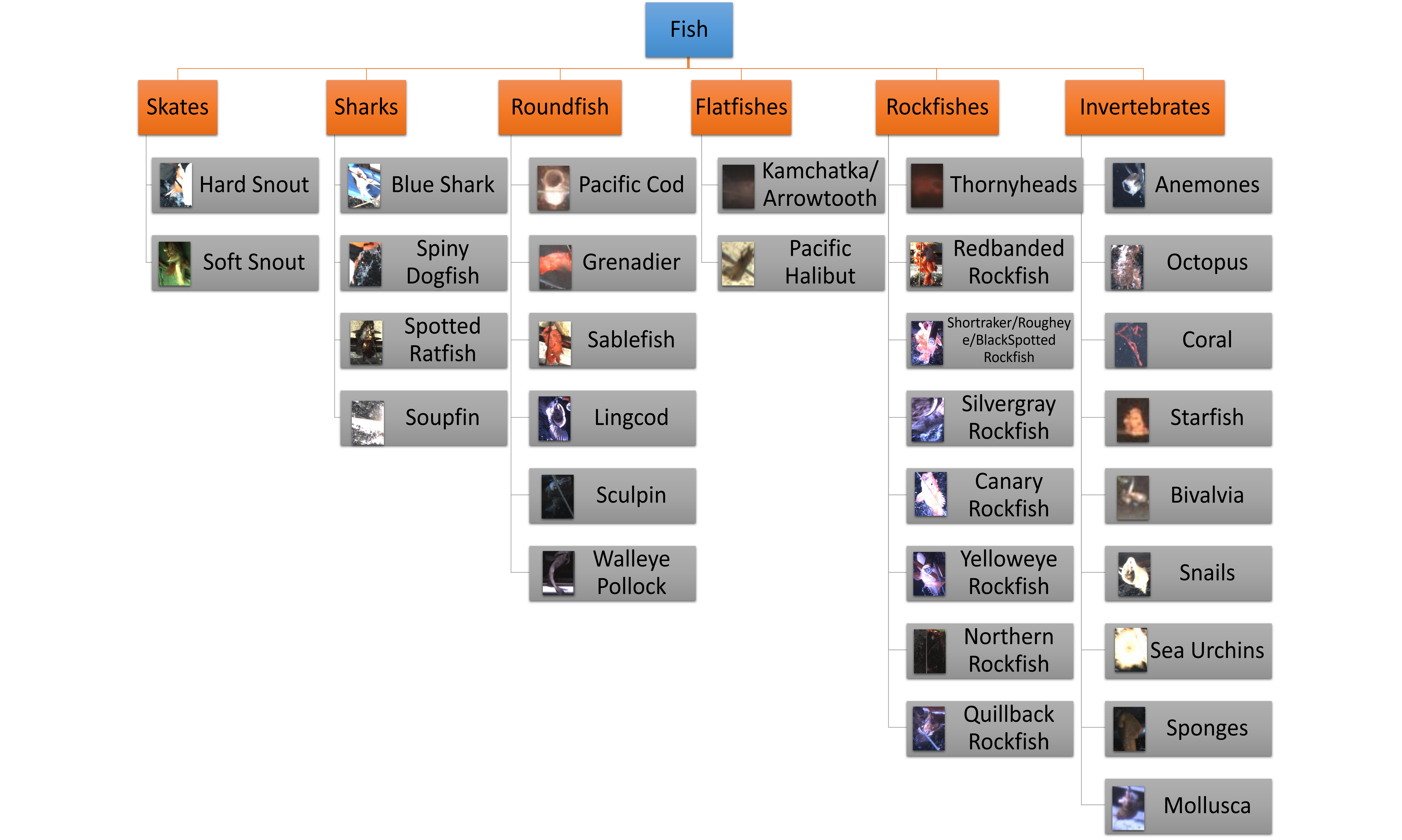}
\caption{Hierarchical Data Structure: The dataset, provided by NOAA fisheries scientists, includes frames and corresponding labels which are bounding box location, start and end frames' IDs of each individual fish, coarse-level group ground truth, and fine-level species ground truth. The sample images shown here are randomly chosen from the dataset.} \label{hierarchical structure}
\end{figure}

\section{\textbf{Related Work}}
\label{Related Work}
\textbf{Hierarchical Classification} The previous work~\cite{Mei2020VideobasedHS} proposes a hierarchical fish species dataset as shown in Fig.~\ref{hierarchical structure}. There are 6 course-level groups and 31 fine-level species. The total number of frames is more than 186K. We also use this dataset to perform class incremental learning. 

The method proposed in work~\cite{Mei2020VideobasedHS} is an end-to-end training approach with a multi-head CNN-based architecture and two levels' loss functions. It outperforms the traditional softmax-based flat CNN classifier in the hierarchical classification task. In contrast to it, our proposed method can not only perform the hierarchical classification task but also maintain good performance on both trained classes and newly acquired classes under the class incremental scenario.

Moreover, work~\cite{Mei2020VideobasedHS} utilizes a video-based inference method, i.e., majority vote, for each individual fish to improve the classification accuracy. For a fair comparison, our proposed HCIL method in this paper, also utilizes the same inference method.

\textbf{Class Incremental Learning} Rehearsal-based methods~\cite{douillard2021dytox, rebuffi2017icarl, hou2019learning, castro2018end} which allow storing a fixed number of data from previously trained classes, have been widely used in the class incremental learning scenarios. More specifically, when training the model on new classes, the model can have access to the stored raw data or feature maps of previously trained classes, which are referred to as memory in the CIL setting. Our proposed HCIL method is also a rehearsal-based method. But contrast to these previous works, which are designed for flat classifiers, our model is a hierarchical classifier that contains a different memory selection module.

Except for the memory, training classes available at the same time are defined as one 'task'. Some incremental learning methods such as~\cite{fernando2017pathnet, serra2018overcoming} require a task identifier at test-time, i.e., need to know which task the test image belongs to. However, our proposed method discards the need for a task identifier by choosing the prediction with the maximum confidence score as the output.

\section{\textbf{Proposed Method}}
\label{Proposed Method}
\begin{figure*}[htpb]
\centering
\includegraphics[width=1\textwidth]{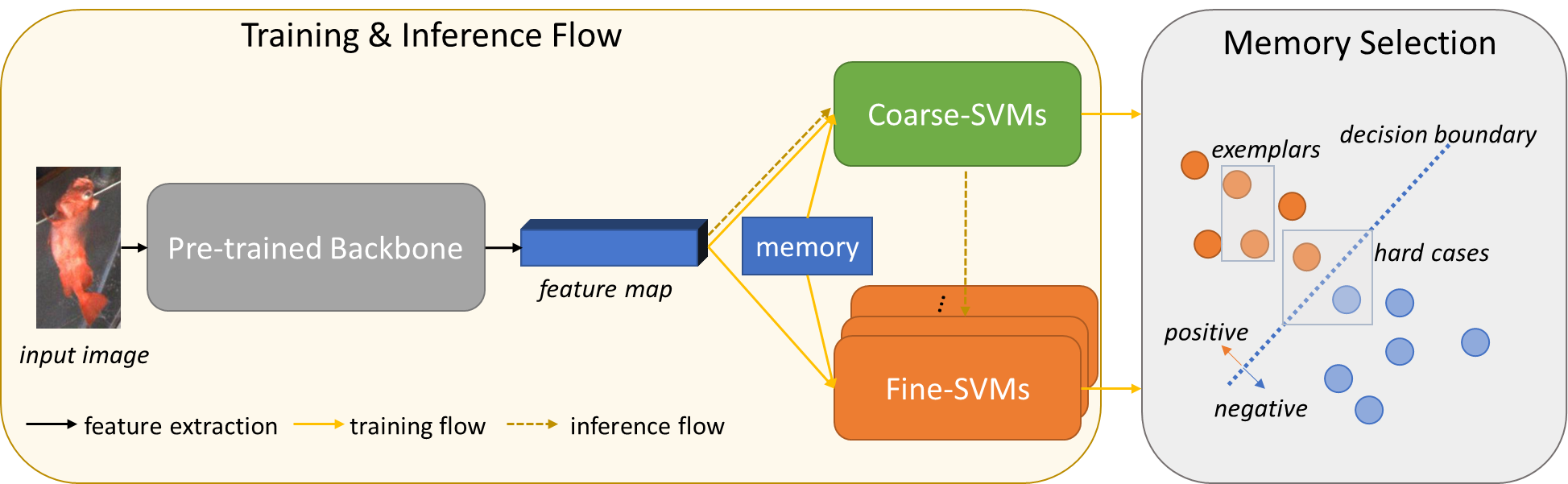}
\caption{HCIL: New classes' feature maps from the fixed pre-trained backbone and old classes' feature maps from CIL memory are used to train coarse-level group SVMs and corresponding fine-level SVMs. After SVMs training, based on the distance between SVMs' decision boundary and each training data, CIL memory keeps adding new classes' features, i.e., hard cases, and adding positive exemplars from new classes based on herding~\cite{welling2009herding}. During inference, the extracted feature map first goes into Coarse-SVMs. And based on the coarse-level group prediction, the extracted feature map goes into corresponding Fine-SVMs for species classification.}
\label{pipeline}
\end{figure*}

Our hierarchical class incremental learning (HCIL) method consists of a fixed pre-trained feature extraction backbone, a CIL memory selection module, and dynamic support vector machines (SVMs), i.e., SVMs are continually added with appearing of newly acquired classes as shown in Fig.~\ref{pipeline}.

\textbf{Pre-trained Feature Extraction Backbone} Due to the easy access to large public datasets such as ImageNet-1k~\cite{russakovsky2015imagenet} and COCO~\cite{lin2014microsoft}, a pre-trained backbone such as ResNet~\cite{he2016deep} or Swin Transformer~\cite{liu2021swin} can be used to extract discriminative features from images even for new datasets or new classes. These public pre-trained backbones can certainly be utilized in our HCIL model.

Despite the remarkable ability of discriminative features extraction of the pre-trained backbone, inevitably it may generate non-discriminative features for some new classes beyond the dataset the backbone pre-trained on. As a result, we propose a CIL memory selection module, to be discussed later, to select those hard cases for the pre-trained backbone. Our method utilizes selected CIL memory to adjust SVM classifiers' boundary, instead of updating the backbone, under the hierarchical class incremental learning scenario.

\textbf{Dynamic SVMs Expansion} During the training of a class incremental learning scenario, the total number of seen classes is increasing. Only the training data for a small number of classes have to be present at the beginning and new classes are added progressively. Thus, the growing-class classifier can be used in the incremental learning models.

Our Coarse-SVMs consist of six SVMs for group-level prediction because there are six groups in the fish dataset as shown in Fig.~\ref{hierarchical structure}. We use one-vs-all training strategy, where for each SVM only one group data are treated as positive and all rest of groups data are treated as negative. Each Coarse-SVM has its own corresponding Fine-SVMs. For example, the Coarse-SVM for the `Sharks' group has four corresponding Fine-SVMs. And these four Fine-SVMs are added into the model sequentially with the availability of newly acquired shark species, which is called `dynamic SVMs expansion' in this paper. During the inference, based on the Coarse-SVMs prediction, the feature map goes to corresponding Fine-SVMs for species classification so that the model can provide hierarchical predictions.

\begin{algorithm}
\caption{Positive Exemplars Selection}
\label{alg:A}
\begin{algorithmic}
\STATE {\textbf{input} feature set $F = \{f_1,...,f_n\}$ of new class $y$}
\STATE {\textbf{input} new class mean, $\mu \leftarrow \frac{1}{n} \sum_{f \in F} f$} 
\STATE {\textbf{input} target number of exemplars, $m_y$} 
\FOR{$k=1,...,m_y$} 
\STATE $p_{k} \leftarrow \underset{f \in F}{\operatorname{argmin}}\left\|\mu-\frac{1}{k}\left[f+\sum_{j=1}^{k-1} p_{j}\right]\right\|$
\ENDFOR
\STATE {\textbf{output} exemplar set $P_y \leftarrow\left(p_{1}, \ldots, p_{m_y}\right)$}
\end{algorithmic}
\end{algorithm}

\textbf{CIL Memory Selection} This module selects some hard cases from newly arrived classes based on their feature maps from the pre-trained backbone. Hard cases is defined by the distances between the feature maps and decision boundary, i.e., the SVM’s outputs, as shown in Fig.~\ref{pipeline}. These distances can be represented as confidence scores, between 0 and 1, from SVM outputs via logistic transformation. The low confidence of a feature map represents a hard case. As incremental learning goes, more and more hard cases are added. In order to fix the total target number $n$ of hard cases, we sort hard cases by their confidence scores and only keep first $n'_i$ low confidence hard cases for each ${SVM}_i$ where $n = \sum_{i} n'_i$.

Besides hard cases selection, the CIL memory module also selects positive exemplars from new classes based on herding~\cite{welling2009herding}. More specifically, positive exemplars are selected based on Algorithm~\ref{alg:A}, where for each new class, exemplars $p_1, . . . , p_{m_y}$ are selected iteratively until reaching the target number, $m_y$. Within each iteration, one more sample of the new class is selected to the exemplar set, namely the one that causes the mean feature vector over all current exemplars to best approximate the mean feature vector over all examples of this new class. Thus, the exemplar set $P_y$, is a prioritized list, i.e., the order of exemplars matters and exemplars earlier in the list are more important. 

 As incremental learning goes, more and more positive exemplars for new classes are added. In order to fix the total number of positive exemplars from all classes, $m$, we lower $m_y$ to $m'_y$ for each class $y$ and do the same thing as done for hard cases, i.e., only keeping the first $m'_y$ exemplars in the exemplar set, $P_y$, for each class $y$.

\section{\textbf{Experimental Results}}
\label{Experimental Results}
We compare our method with state-of-the-art work~\cite{Mei2020VideobasedHS} on NOAA's dataset under both `hierarchical classification setting' and `hierarchical class incremental learning setting'. 

\textbf{Hierarchical Classification Setting} This setting serves as a baseline for the next hierarchical class incremental learning setting. More specifically, in this setting, all training data for all classes are available at the same time, i.e., no incremental learning scenario is assumed. Training and testing data split is the same as~\cite{Mei2020VideobasedHS}, where each individual fish has its own video data. Training and testing fish are totally different individual fish.

In this setting, our proposed \textbf{HCIL} model directly uses the fixed ResNet-101 backbone pre-trained on ImageNet-1k as the feature extractor and only trains our SVMs on NOAA's dataset. No CIL memory is involved, thus noted as \textbf{`HCIL w/o m'} in Table~\ref{HC}. As mentioned in Section~\ref{Proposed Method}, during inference, based on the Coarse-SVMs prediction, the feature map goes to corresponding Fine-SVMs for species classification so that the model can provide hierarchical predictions. 

For a fair comparison, we allow work~\cite{Mei2020VideobasedHS} to utilize the same pre-trained ResNet-101 as the initial backbone, which is also further finetuned along with its classifier heads on NOAA's hierarchical fish dataset. In Table~\ref{HC}, we report image-based accuracy and video-based accuracy, noted as $img$ and $video$ respectively, on both coarse (group) level and fine (species) level, noted as subscript $C$ and $F$ respectively. 

Results are in Table.~\ref{HC}. Even though our proposed \textbf{`HCIL w/o m'} method fixes the pre-trained backbone, which is not finetuned by the NOAA's fish dataset, the performance is still comparable with work~\cite{Mei2020VideobasedHS}, that allows updating the backbone and calculating cross-entropy loss functions on both levels. This shows the backbone pre-trained on large public datasets without finetuning can indeed possess the strong ability to extract discriminative features even on new classes or datasets.

\begin{table}[htbp]
    \caption{Hierarchical Classification Setting}\label{HC} \centering
    \setlength{\tabcolsep}{1.5mm}{\begin{tabular}{cccccc}
        \toprule[1.5pt]
        Method & $img_C$ & $img_F$ & $video_C$ & $video_F$ \\
        \midrule[1pt]
        ~\cite{Mei2020VideobasedHS} & $92.0$ & $82.9$ & $96.5$ & $91.2$\\
        HCIL w/o m & $91.8$ & $81.2$ & $95.9$ & $90.7$\\
        \bottomrule[1.5pt]
    \end{tabular}}
\end{table}

\begin{table}[htbp]
    \caption{Hierarchical Class Incremental Learning Setting}\label{HCIL} \centering
    \setlength{\tabcolsep}{1.5mm}{\begin{tabular}{cccccc}
        \toprule[1.5pt]
        Method & $img_C$ & $img_F$ & $video_C$ & $video_F$\\
        \midrule[1pt]
        ~\cite{Mei2020VideobasedHS} w/ m & $80.5$ & $65.7$ & $81.1$ & $70.4$\\
        HCIL w/o m & $82.8$ & $69.9$ & $86.2$ & $75.3$\\
        HCIL       & $91.0$ & $80.4$ & $92.1$ &$82.8$\\
        HCIL w/ Swin  & $\mathbf{92.6}$ & $\mathbf{83.5}$ & $\mathbf{93.2}$ & $\mathbf{84.3}$\\
        \bottomrule[1.5pt]
    \end{tabular}}
\end{table}

\textbf{Hierarchical Class Incremental learning Setting} In this setting, both our method, HCIL, and work~\cite{Mei2020VideobasedHS} still utilize ResNet-101~\cite{he2016deep} backbone pre-trained on ImageNet-1k. Testing data are still the same as the previous setting and cover all species. However, training data are divided into three \textbf{tasks}. The first task includes one-third of the species within each group. The second task includes another one-third of the species within each group. The third task includes the rest species' data. There are no overlapping classes between the three tasks. However, the `Skates' and `Flatfish' groups have only two species each so one Fine-SVM for each group is enough. As a result, all data from these two groups are included in the first task. 

When training our HCIL model, on the first task data, the feature maps from the fixed pre-trained backbone are used to train six Coarse-SVMs and corresponding Fine-SVMs. Based on SVMs confidence scores and the herding method introduced in Section~\ref{Proposed Method}, HCIL constructs the memory. Next, when training on the following tasks, the memory's feature maps are also used to train Coarse-SVMs and newly added Fine-SVMs. We set the total number of hard cases $n$ to $200$, and the total number of positive exemplars $m$ to $1800$ so that the memory size won't increase as the incremental learning goes.

Work~\cite{Mei2020VideobasedHS} does not have memory selection or classifiers expansion. For a fair comparison, when finetuning work~\cite{Mei2020VideobasedHS}'s both pre-trained backbone and classifiers on later two tasks, we allow it to use the same size memory but randomly sampled from previously trained classes, denoted as \textbf{`~\cite{Mei2020VideobasedHS} w/ m'} in Table~\ref{HCIL}. When training on each task, its classifiers always output predictions over 31 species and calculate loss functions. We evaluate the final trained models on all testing data. 

Results are in Table.~\ref{HCIL}. Compared with work~\cite{Mei2020VideobasedHS} which is not designed for incremental learning, our HCIL model achieves significantly better performance. For the CIL memory ablation study, we remove CIL memory, noted as \textbf{`HCIL w/o m'} in Table~\ref{HCIL}, and the performance drops dramatically but is still better than work~\cite{Mei2020VideobasedHS} with randomly sampled memory. This ablation study shows the benefits of CIL memory selection module. It also tells that under incremental learning setting, updating the backbone and classifiers even with some randomly sampled memory, makes the deep model suffer from catastrophic forgetting. For the backbone ablation study, we replace ResNet-101~\cite{he2016deep} with Swin-Transformer~\cite{liu2021swin}, noted as \textbf{`HCIL w/ Swin'}, and get the best performance.

\section{\textbf{Conclusions and Future}}
\label{Conclusions}
Our proposed HCIL model combines the advantages of both hierarchical classification and incremental learning. It consists of a fixed backbone pre-trained on large public datasets, a CIL memory selection module, and dynamic SVMs expansion. Our HCIL is also a backbone-agnostic approach. Future experiments may include dividing training data into more tasks to form longer incremental learning scenarios.

\bibliographystyle{IEEEbib}
\bibliography{strings,refs}

\end{document}